\documentclass[runningheads]{llncs}
\usepackage[T1]{fontenc}
\usepackage{amsmath,amssymb,amsfonts}
\usepackage{algorithm}
\usepackage{algorithmic}
\usepackage{pythonhighlight}
\usepackage{comment}
\usepackage{subcaption}
\usepackage{graphicx}
\usepackage[numbers]{natbib}
\definecolor{commentcolor}{RGB}{110,154,155}   
\newcommand{\PyComment}[1]{\small\ttfamily\textcolor{commentcolor}{\# #1}}
\newcommand{\PyCommentident}[1]{\small\ttfamily\textcolor{commentcolor}{\hspace{8pt}\# #1}}
\newcommand{\PyCode}[1]{\small\ttfamily\textcolor{black}{#1}} 
\newcommand{\PyCodeB}[1]{\ttfamily\textcolor{black}{#1}} 
\newcommand{\frd}[1]{\hspace{1pt}} 

\begin{document}
\title{Point-GR: Graph Residual Point Cloud Network for 3D Object Classification and Segmentation}

\author{Md Meraz\inst{1} \inst{3}\and
Md Afzal Ansari \inst{1} \inst{3}\and
Mohammed  Javed\inst{2} \inst{3}\and
Pavan Chakraborty \inst{1} \inst{3}
}

\institute{
Center of Intelligent Robotics Lab, Department of Information Technology \\
\and
Computer Vision \& Biometrics Lab, Department of Information Technology \\
\and
Indian Institute of Information Technology Allahabad, Prayagraj-U.P, India. \\
Corresponding author: Md Meraz \email{pro2016001@iiita.ac.in} 
}
\authorrunning{M. Meraz  et al.} \maketitle              

\begin{abstract}
In recent years, the challenge of 3D shape analysis within point cloud data has gathered significant attention in computer vision. Addressing the complexities of effective 3D information representation and meaningful feature extraction for classification tasks remains crucial. This paper presents Point-GR, a novel deep learning architecture designed explicitly to transform unordered raw point clouds into higher dimensions while preserving local geometric features. It introduces residual-based learning within the network to mitigate the point permutation issues in point cloud data. The proposed Point-GR network significantly reduced the number of network parameters in Classification and Part-Segmentation compared to baseline graph-based networks. Notably, the Point-GR model achieves a state-of-the-art scene segmentation mean IoU of 73.47\% on the S3DIS benchmark dataset, showcasing its effectiveness. Furthermore, the model shows competitive results in Classification and Part-Segmentation tasks.

\end{abstract}

\section{Introduction}

The advancement of computer vision and artificial intelligence tools has led to the widespread use of autonomous vehicles, industrial robots, and other technologies for the convenience of society. Developing a sophisticated and precise perception system that emulates human oriented actions is crucial but challenging for a robot. Humans are capable of performing tasks such as 3D Object Classification, Recognition, and 3D Segmentation with precision in their day-to-day routines. However, for autonomous system, capturing and processing 3D world information poses difficulties due to the limitations in range capturing devices and 3D data processing software and compute hardware. This research paper aims to explore the use of point cloud data to efficiently deal with challenging tasks, specifically focusing on Object Classification, Part-Segmentation and Scene Segmentation.

Object Classification and Segmentation on point cloud datasets can be approached through handcrafted features or data-driven methods. Previous studies \cite{sun2009concise, rusu2008aligning, johnson1999using} have contributed significantly to the field using handcrafted features. Nevertheless, data-driven methods have recently risen to prominence because of their proficiency in local as well as global features learning from point cloud data.These approaches can be categorized as voxel-based representation \cite{wu20153d}, point-based \cite{qi2017pointnet, qi2017pointnet++}, and graph-based \cite{Simonovsky_2017_CVPR, xu2020grid, wang2019dynamic} method.

Voxel-based technique convert point cloud data into regular structures, such as grids or voxels, which are then used as input for 3D Convolutional Neural Networks(CNNs) for Classification. Maturuna et al. \cite{maturana2015voxnet} introduced volumetric occupancy networks, which have a limitation with respect to time and space, determined by voxel size. Klokov et al. \cite{Klokov_2017_ICCV} proposed KD-tree to reduce the space complexity of volumetric networks. Other advanced techniques, like Octree-based networks \cite{riegler2017octnet}, have further improved space and time complexities. Dense Point \cite{iandola2014densenet}, Interp CNN \cite{mao2019interpolated}, and Geo CNN \cite{lan2019modeling} also utilize voxel-based approaches to extract point cloud features.

Point-based methods PointNet \cite{qi2017pointnet} was the first influential model that directly uses 3D point cloud data. PointNet ensures permutation in-variance by employing a symmetric function and global features are aggregated through a MaxPooling layer. PointNet$++$ \cite{qi2017pointnet++} extends PointNet by introducing a hierarchical network for each sampling point group. 

Graph based Methods have gained popularity for handling unstructured data using graph algorithms \cite{Simonovsky_2017_CVPR}. Graph-based models, unlike GNN-based models, represent sets of point cloud as a graph. A typical point cloud graph represents  with nodes and edges, further perform convolution neural network(CNN) and pooling operations on it. Simonovsky et al. \cite{Simonovsky_2017_CVPR} connect all 3D points in a point cloud as a nodes and utilize edge based convolution with max-pooling operation to accumulate neighborhood information. Graph Neural Networks (GNN) \cite{xu2020grid} have also been utilized for point cloud data classification. This work is motivated by the supreme features of PointNet \cite{qi2017pointnet} and DG-CNN \cite{wang2019dynamic}, combining permutation in-variance, global feature learning, and dynamic graph construction.

Some research works have extended object classification networks to achieve scene segmentation and part segmentation. The primary contributions are described in the present research work is listed below:
   \begin{itemize}
      \item 
      A cutting-edge deep learning based model Point-GR, which directly transform unordered sets of raw point cloud into higher dimension using Point-GR Transformation network as well as preserves local geometric feature.
      \item 
      Introducing the residual based learning in the Point transformation to solve point permutation for point cloud data.
      \item The proposed network reduced the number of network parameters in classification and Part-Segmentation model together achieves state-of-the-art Scene Segmentation performance(Mean-IoU: 73.47\%) on S3DIS benchmark dataset. 
      \item The Point-GR model demonstrates comparable performance in both Classification on the ModelNet-40 dataset and Part-Segmentation on the ShapeNet-Part dataset.
    \end{itemize}
     
The whole research paper is divided into four major sections. The $2^{nd}$ Section discusses the proposed model, $3^{rd}$ Section reports on detailed experimental results and discussions with ablation studies, and $4^{th}$ Section summarizes the research work.

\section{Proposed Methodology}
The research project focuses on a point-based method for object classification and segmentation using sparse, variable LiDAR-generated point clouds. Each LiDAR sample, with $N$ points, can be represented in $N!$ combinations, requiring models to be invariant to point permutations. This issue was first addressed by the PointNet \cite{qi2017pointnet} architecture.


\begin{figure}[!b]
    \centering
    {\includegraphics[width=12cm]{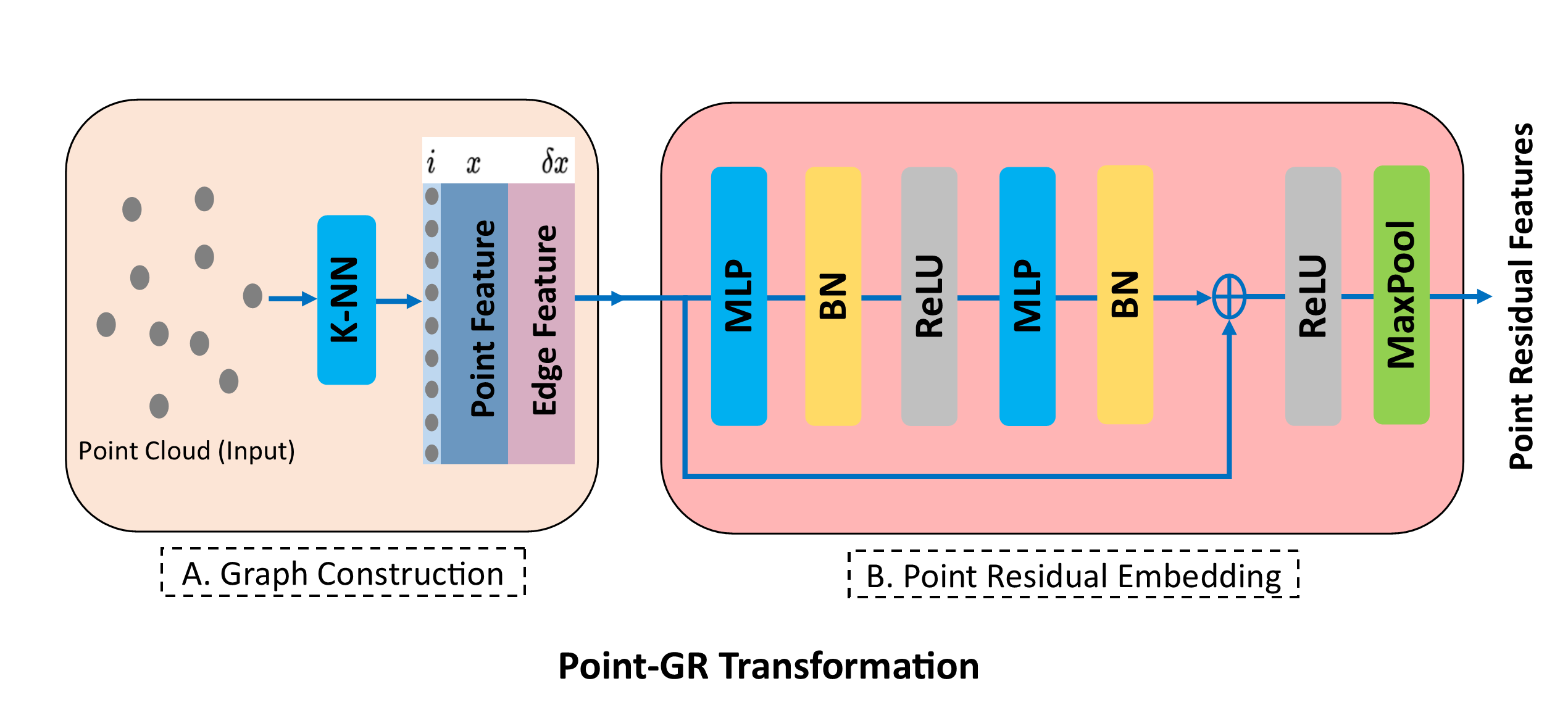} }
    	\caption{The figure shows the Point-GR Transformation network. It consists of Graph Construction and Point Residual Embedding blocks.}
    \label{Point GR Transform}
\end{figure}

In the proposed research work, we have introduced a Point-GR Transformation network which allows to capture the complex features while mitigating with the point in-variance, sparsity and noise. The proposed architecture consists of two main block Point-GR Transformation and Point-GR Feature Learning Network.
Each block has graph construction from a point cloud data with a learning neural network depending on the objective. It is individually examined and discussed in the following sections providing a comprehensive model overview.

\subsection{Point-GR Transformation}

The objective of the Point-GR Transformation module is to transform the point cloud(N x 3) data into a high dimensional space and then revert to lower dimension. In this process the model represent the point cloud information into a concise(robust) representation. It consist of two main parts (shown in Fig. \ref{Point GR Transform}) i.e. Graph construction and Point Residual Embedding which are discussed in the following subsections.

\subsubsection{Graph Construction}
The initial phase in a graph-based convolution neural network(GNN) involves the pre-processing of the point cloud data, followed by the construction of a graph. A raw point cloud data is represented as a set of 3D points $P_i$ :: $i = 1,2,3,4,5...... N $, 
where each point is minimum $(x, y, z)$ values with additional information like laser peak, width, area, radial-distance.
At first stage, a graph is constructed using a $k$-nearest neighbors $k$-NN method \cite{altman1992introduction}. This approach groups the points that are in closest proximity to point $P_i$, creating a localized graph. 

For a given point cloud, denoted as $P$, we have generated a point graph $G = (P, \hspace{2pt} E)$ in which the points serve as a vertices or local point feature and E is termed as graph edge also edge features. The overall graph is formed by connecting each point to its closest neighbors through the $k$-nearest neighbors ($k$-NN) method. The value of  k is decided by experiment in general these are the possible value k- 5/10/20/40 like \cite{dc-gnn}.

    
    


The 3D point cloud edge features is calculated by euclidean distance method. 
\vspace{-1pt}
\begin{flalign}
 d \hspace{2pt} = \hspace{2pt} \{(P_i, P_j) \hspace{2pt}\mid \hspace{2pt} min_k({\mid\mid P_i - P_j \mid\mid}_2) \}
\end{flalign}

 Corresponding to each points, denoted as $p_i$, a feature vector is generated by integrating both edge feature and point feature, as specified by the equations-(2,3).
    \begin{flalign}
    d_p \hspace{2pt} = \hspace{2pt} P_i,\hspace{2pt} d_e\hspace{2pt} = \hspace{2pt} P_i - P_j
    \end{flalign}
    
    \begin{flalign}
    d_{pe}{\{i=1.....N\}} \hspace{2pt} = \hspace{2pt} [d_p,\hspace{2pt}d_e] 
    \end{flalign}
    
    where $P_i$ $\in$ $\{x_i,y_i,z_i \}$, $d_{pe}$ $i^{th}$ is feature vector, $d_p$ is point feature vector, $d_e$ is edge feature vector.

\subsubsection{Point Residual Embedding}
We have introduced a Point Residual Embedding block in the network which transforms point cloud data into a high level representation. 
The raw point clouds are converted into canonical structures i.e. graphs as mentioned in the previous sub-section. Further the convolutions are performed on the point cloud graph structures. The feature vectors of each vertices are exchanged between connected vertices via their edges. When considering a point, let's say, $P_i$ belonging to the graph G, convolutions are conducted with respect to this reference point by exchanging the feature or state vectors along with updating them concurrently. The resulting updated feature vector 
passed through convolution, batch normalization and ReLU activation function which further passed to the second convolution layer. The output from second convolution is again passed through the batch normalization only.

Following the convolution step, skip connections are established, where the output from the previous layer is added to the output of the convolution layer.  This approach eliminate the issue of vanishing gradient in the network, and the output is passed through ReLU activation function. The final output produced by the network is selected depends on max value of local k-nearest local graph.



\begin{figure}[!t]
    \centering
    {\includegraphics[width= 12cm]{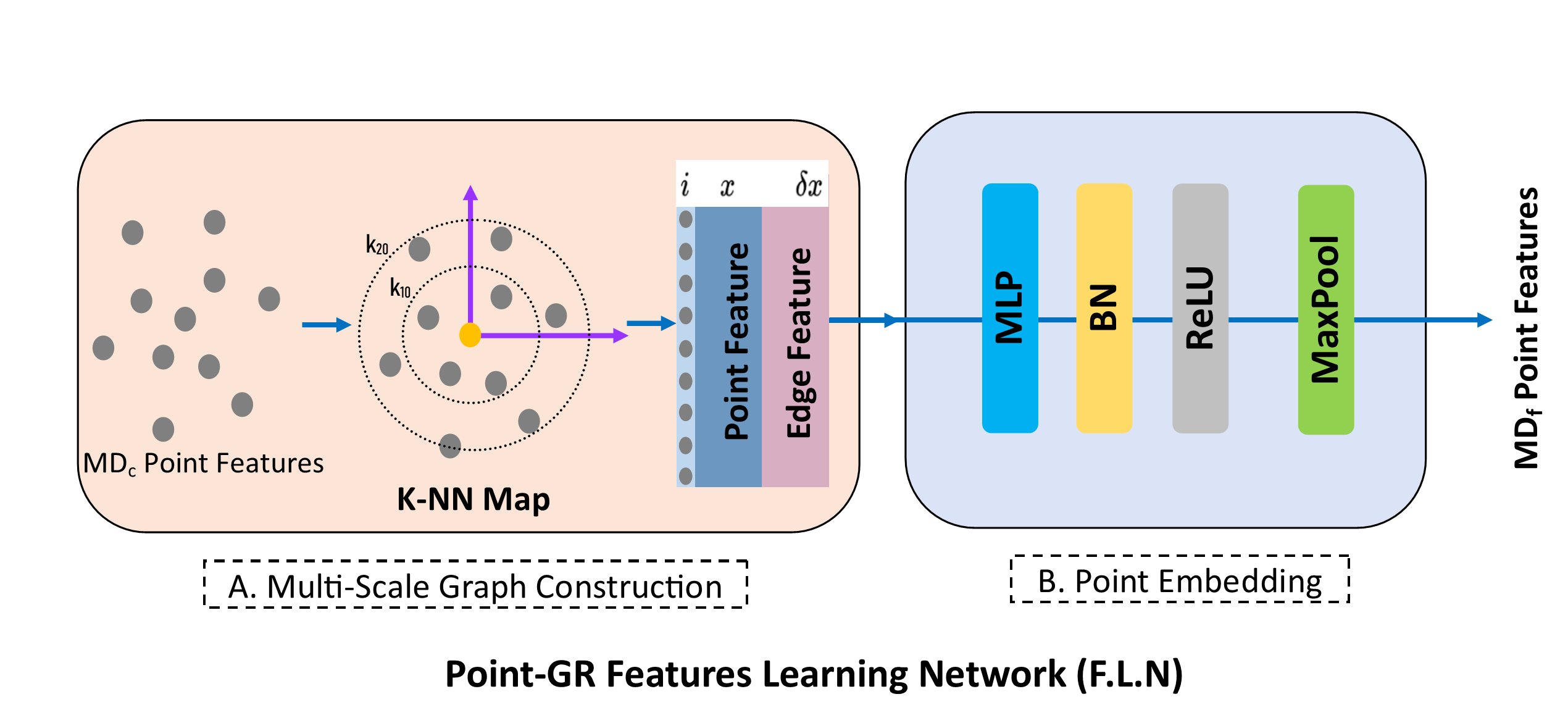} }
    	\caption{Shows the Point-GR Features Learning Network(F.L.N). It consists of Multi-Scale Graph Construction and Point Embedding block.}
    \label{FLN}
\end{figure}

\subsection{Point-GR Feature Learning}

 The output of Point-GR Transformation network i.e. point residual feature(PRF) is passed through the Point-GR Feature Learning Network(FLN). It consists of two main blocks Multi-scale Graph Construction and Point Embedding Fig. \ref{FLN}. This block is used multiple times in the Point-GR network.

Multi-Scale Graph construction is similar to Point-GR Graph construction as shown in Fig. \ref{Point GR Transform} but it also capable of handling dynamic channel point cloud as a input. Point Embedding network is similar to the ResNet like feature extraction network except without the skip connections. In this block output of Multi-scale Graph Construction network is taken as input. In the first stage the input is passes to MLP, batch normalization with ReLU activation function. Further it is passed through a max-pooling layer to generate the final high dimensional output vector $MD_f$ point features. For hierarchical learning $MD_f$ Point Features acts as input($MD_C$ Point Features) of next $FLN$ block in Fig. \ref{FLN}.

\begin{figure}[!b]
    \begin{subfigure}{\textwidth}
      \centering
      \includegraphics[width=12cm]{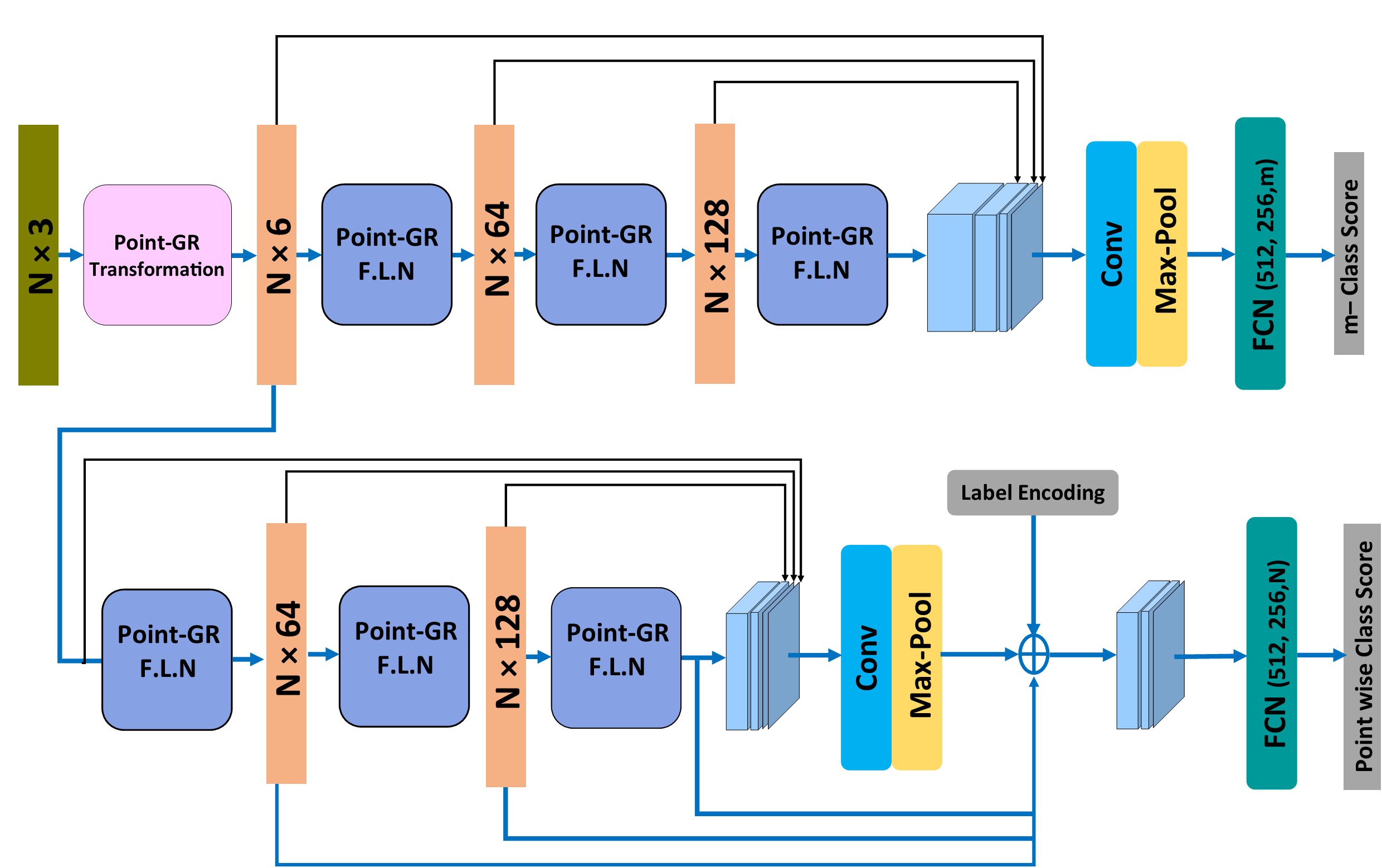}  
    \end{subfigure}

    \caption{Deep learning based architecture of (a) The extended Point-GR architecture for Classification Network which includes transformation networks, and it generates $m$-class scores for point cloud input. and (b) The extended Point-GR architecture for Part-Segmentation which includes transformation networks, and it generates $N\times c$ segmentation scores for point cloud data.}
      \label{class-seg}
\end{figure}
\subsection{Point-GR Classification and Segmentation Architecture}


The Point-GR architecture, depicted in Fig. \ref{class-seg}, constitutes a classification network designed for point cloud data and the detailed description of classification algorithm is given in pseudo code form in Algorithm \ref{algo2}. At initial phase involves transforming the input into Point Residual Features (PRF) using the Point-GR Transformation Network. 
The Point-GR Transformation Network consists of two convolution layer first the input dimension is transformed into higher dimension 64 and then second convolution layer squeezes dimension into 6, the final output of Point-GR Transformation is $[N \times 6]$.

The Point-GR Transformation network is fed to the Feature Learning Network. The Feature Learning Networks combines point and edge features to formulate the Point features. There are three sequential FLN blocks present in the network which performs the conv2D operations on it. These phases exhibit progressive embedding dimensions represented as $MD_f$ = $\{6, 64, 128, 256\}$, alongside the sampled point count.Skip connections from each phase are established and concatenated, forming a unified feature vector that encapsulates multi-scale features. The final feature vector is processed through a layer of convolution with max-pooling. At the end of the Point-GR classification network, a fully connected network is applied to map the feature vector $(512, 256, m)$ to m-classes score where m denotes the classes in a dataset.

\begin{algorithm}[!t]
\PyCodeB{Input:} \\
\PyCode{P - }  \PyCode{[N, 3], Input coordinates of raw point cloud, where N denotes the total number of points.} \\ 
\PyCode{k - number of the neighboring points in k-NN algorithm.} \\
\PyCode{f - number of kernel, } \hspace{4pt} {label - number of categories.}\\
\PyCode{LFE - Label Feature Encoding.} \hspace{4pt}{FV - Feature Vector} \\
\PyCode{Output:} \\
\PyCode{class\_score - classification score of point cloud.} \\
\PyCode{segmentation\_score - Point-wise segmentation score.} \\

\PyComment{Point-GR Network for classification} \\
\PyCode{Function Classification(P, k)}  \\    
\PyCode{\hspace{4pt} MS\_c = PRF = Point-GR Transformation(P, k)} \\
\PyCommentident{sequential feature encoding MS\_{f1}...MS\_{f3}}\\
\PyCode{\hspace{4pt} MD\_f1 = FLN(MD\_c, f, k)} \\
\PyCode{\hspace{4pt} MD\_f2 = FLN(MD\_f1, f, k)} \\
\PyCode{\hspace{4pt} MD\_f3 = FLN(MD\_f2, f, k)} \\
\PyCode{\hspace{4pt} MD\_{out} = concat(MS\_c, MD\_f1, MD\_f2, MD\_f3)} \\
\PyCode{\hspace{4pt} MD\_{out} = conv\_layer(MD\_{out})} \\
\PyCode{\hspace{4pt} return FCN(MD\_{out}.flatten)} \\

\PyComment{Point-GR Network for Part Segmentation} \\
\PyCode{Function PartSegmentation(P, k, label)}  \\
\PyCode{\hspace{4pt} MS\_c = PRF = Point-GR Transformation(P, k)} \\
\PyCommentident{sequential feature encoding MS\_{f1}...MS\_{f3}}\\
\PyCode{\hspace{4pt} MD\_f1 = FLN(MD\_c, f, k)} \\
\PyCode{\hspace{4pt} MD\_f2 = FLN(MD\_f1, f, k)} \\
\PyCode{\hspace{4pt} MD\_f3 = FLN(MD\_f2, f, k)} \\
\PyCode{\hspace{4pt} MD\_{f} = concat(MS\_c, MD\_f1, MD\_f2, MD\_f3)} \\

\PyCode{\hspace{4pt} Features\_{out} = conv\_layer(MD\_{f})} \\
\PyCode{\hspace{4pt} {Label\_Encoder} = Linear(label)} \\
\PyCode{\hspace{4pt} LFE = concat(Label\_Encoder, Features\_{out})} \\
\PyCode{\hspace{4pt} {FV} = concat(LFE, MS\_c, MD\_f1, MD\_f2, MD\_f3)} \\
\PyCode{\hspace{4pt} return FCN(FV)} \\

\PyComment{In Each Processing Stage:}\\
\PyCode{ class\_score = Classfication(P, k)}\\
\PyCode{ segmentation\_score = PartSegmentation(P, k, label)}
\caption{Psuedo Code of the PointGR: Classification, Part and Indoor Segmentation}
\label{algo2}
\end{algorithm}



Second model, Part Segmentation network comprises of two parts. In the first part the raw point clouds are passed through the Point-GR Transformation layer which encapsulates the input into PRF(Point Residual features) similar to Point-GR classification Network as shown in lower section of Fig. \ref{class-seg}. The second part includes Future Learning Network(FLN) which is similar to classification network and feature increases with dimension $MD_f = \{6, 64, 128, 256\}$. It is further passed through the conv1D operation and a max-pooling layer. The skip connections from the embedding layer and the other two graph networks, along with the labels encoding are combined to form a new point features. A fully connected network is present at the last layer of the network to predict the part segmentation scores for the $N$ point cloud data. Similarly the Point-GR architecture for segmentation task is also applicable for indoor scene segmentation. 

\section{Experimental Results}
Current section includes experimental results which are organized into four subsections, each providing a detailed description of the dataset, model configuration, loss function, optimizer, accuracy and hyperparameters of the networks. The initial subsection presents and compares object classification accuracy. Following that, the second subsection showcases 3D part-segmentation results. Similarly, the third subsection demonstrates the state-of-the-art 3D points segmentation, and in the fourth subsection, we conduct experiments to configure parameters, studying model robustness. All experiments presented herein conducted on linux platform(Ubuntu 22.04) and open-source package, specifically Python 3.9.13, PyTorch 2.0, and CUDA 12.2 Nvidia GPU.

\subsection{3D Object Classification}
The task of 3D object classification is benchmarks using the ModelNet-40 dataset \cite{wu20153d}. The selection of 3D CAD models representing 40 distinct classes was informed by the statistical insights derived from the SUN database. This dataset comprises a comprehensive collection, including a total of 12,311 3D-CAD models spanning the 40 specified object categories. In alignment with established conventions, our research employs standard train/test(validation) splits\cite{qi2017pointnet}, allocating 9,843 samples for training and 2,468 samples for rigorous evaluation purpose.

\begin{table}[]
    \caption{ 
    Evaluation of 3D object classification results performed on the ModelNet-40 dataset. In this context, P represents the total number of points fixed to 1,024, and N denotes surface normal.}
    \label{table:1}
	\begin{center}
    \begin{tabular}{l|l|r|r}
      \hline \textbf{Method} & \textbf{input} & \textbf{Points} & \textbf{Accuracy} \\
      \hline 
      PointNet \cite{qi2017pointnet} & P & $1 \mathrm{K }$ & $89.2 \%$ \\
      SO-Net \cite{Li2018SONetSN}  & P, N & $2 \mathrm{K}$ & $90.9 \%$ \\
      A-CNN  \cite{A-CNN} & P, N & $1 \mathrm{K}$ & $92.6 \%$ \\
      PointNet++ \cite{qi2017pointnet++}  & P & $1 \mathrm{K}$ & $90.7 \%$ \\
      P2Sequence  \cite{guo2020pct} & P & $1 \mathrm{K}$ & $92.6 \%$ \\
      Kd-Net \cite{kdnet} & P & $32 \mathrm{K}$ & $91.8 \%$ \\
      PCNN  \cite{pcnn} & P & $1 \mathrm{K}$ & $92.3 \%$ \\
      PointCNN  \cite{Li2018PointCNNCO} & P & $1 \mathrm{K}$ & $92.5 \%$ \\
      PointConv \cite{Wu2019PointConvDC} & P, N & $1 \mathrm{K}$ & $92.5 \%$ \\
      RS-CNN  \cite{liu2019relation}& P & $1 \mathrm{K}$ & $92.9 \%$ \\
      
      KPConv \cite{kpconv}  & P & $7 \mathrm{K}$ & $92.9 \%$ \\
      DG-CNN \cite{wang2019dynamic} & P & $1 \mathrm{K}$ & $92.9 \%$ \\
      PointGrid \cite{Le2018PointGridAD}  & P & $1 \mathrm{K}$ & $92.0 \%$ \\
      
      DC-GNN \cite{dc-gnn}  & P & $1 \mathrm{K}$ & $93. 64 \%$ \\
      \hline
     \textbf{Ours} & P & $1 \mathrm{K}$ & $\mathbf{92 . 71 \%}$ \\
     \hline
   \end{tabular}
   \end{center}
\end{table}

\textbf{Training:} For the experiment of Point-GR classification architecture, we process standard 1,024 number of uniform sample points and fed as a input to the network. At each stage of the network local feature dimension is set to 20(i.e. k nearest neighbor points) maintaining during the entire training process. During the Point-GR Transformation stage, we transformed the input points cloud into a higher number of features in the same dimension, effectively doubling it. To compute the network losses, we employed categorical-cross-entropy loss, and for the network optimization, we chose the stochastic gradient descent(SGD) technique with learning rate of 0.1. In contrast, for improved optimization, we also applied learning rate scheduler during training process and momentum was set to 0.9. The end-to-end Point-GR architecture on a batch of 32 point cloud sample. The entire training happened on Nvidia Tesla-V100 GPU.

\textbf{Results:}
The results of object classification on the ModelNet-40 dataset are presented in Table- \ref{table:1}. Our model demonstrates average performance with an accuracy of 92.71\%, surpassing most of the state-of-the-art networks. Compared to the baseline DG-CNN model, our network achieves a 0.21\% decrease in accuracy.


\subsection{Part-Segmentation}      
    Here, we evaluate the performance of our extended Point-GR architecture using the ShapeNet-Part benchmark dataset \cite{yi2016scalable} for part segmentation. The ShapeNet-Part dataset comprises a total of 16,880 3D models spanning 16 object categories, each annotated with over 50 parts. Each sample includes 2-6 different parts. For consistency with previous work \cite{qi2017pointnet}, we adopt the same train-test split standard during experiments.

\begin{table*} 
	\begin{center}
	\begin{small}
	\setlength{\tabcolsep}{0.000001em}
	\begin{center}
	\caption{Evaluation of standard Part-Segmentation model with respect to metric IoU and object IoU on ShapeNet part benchmark dataset(total number of points is 2,048).Here, bold characters signifies highest performance in a column.\\}
    \label{part-seg-results}

	\end{center}       
    \begin{tabular}{l|c|cccccccccccccc}

      \hline 

        \text{Method} &  \text{m-}  &  \text{mug} &  \text{bag}   &  \text{motor}  &  \text{cap} &  \text{knife}  &  \text{pistol}  &  \text{ear}  &  \text{roc-}  &  \text{lamp} &  \text{lap} &  \text{air-}   &  \text{car} &  \text{gui-} &  \text{table}   \\
        & IoU&   &  &  &  &  & & pho & ket&&tar& plane &&tar\\
        
        \textbf{\# Shapes}  &  & 184 & 76 & 202 & 55  & 392 & 283  & 69   & 66   & 1547 & 451 & 2690   & 898 & 787 & 5271    \\ \hline
        
        Kd-Net  & 82.3   & 86.7   & 74.6  & 57.4   & 74.3   & 87.2   & 78.1 & 73.5 & 51.8 & 81.0 & 94.9   & 80.1   & 70.3   & 90.2   & 80.3   \\ \hline
        PointNet   & 83.7   & 93.0   & 78.7  & 65.2   & 82.5   & 85.9   & 81.2 & 73.0 & 57.9 & 80.8 & 95.3   & 83.4   & 74.9   & 91.5   & 80.6    \\ \hline
        SO-Net   & 84.9   & 94.2   & 77.8  & 69.1   & 88.0   & 83.9   & 80.9 & 73.5 & 53.1 & 82.8 & 94.8   & 82.8   & 77.3   & 90.7   &  \textbf{83.0} \\ \hline
        PointNet++  & 85.1   & 94.1   & 79.0  & 71.6   & 87.7   & 85.9   & 81.3 & 71.8 & 58.7 & 83.7 & 95.3   & 82.4   & 77.3   & 91.0   & 82.6   \\ \hline
        PCNN  & 85.1   & 94.8   & 80.1  &  \textbf{73.2} & 85.5   & 86.0   & 83.3 & 73.2 & 51.0 & 85.0 & 95.7   & 82.4   & 79.5   & 91.3   & 81.8   \\ \hline
        DG-CNN    & 85.2   & 94.9   &  83.4 & 66.3   & 86.7   & 87.5   & 81.1 & 74.7 &  \textbf{63.5} & 82.8 & 95.7   & 84.0   & 77.8   & 91.2   & 82.6   \\ \hline
        P2Sequence  & 85.2   & 94.6   & 81.8  & 70.8   &  \textbf{87.5} & 86.9   & 79.3 & 77.1 & 58.1 & 83.9 & 95.7   & 82.6   & 77.3   & 91.1   & 82.8   \\ \hline
        DC-GNN    & 84.5  & 95.2   & 82.2 & 72.8   & 86.0  &  \textbf{88.0} &  \textbf{84.53} & 78.9 & 54.28   &  \textbf{85.02} &  \textbf{95.8} &  \textbf{85.3} &  \textbf{81.0} &  \textbf{92.2} & 82.9 \\ \hline
        PointConv  &  \textbf{85.7} & -   & -  & -   & -   & -   & - & - & - & - & -   & -   & -   & -   & -   \\ \hline
        \textbf{Ours} & 85.2   &  \textbf{95.2} &  \textbf{83.7}  & 72.7  & 84.0  & 86.85  & 82.62   &  \textbf{79.4} &  63.02 & 83.5   & 95.6  & 84.7  & 79.8  & 91.5  & 82.3    \\ \hline
     
      \end{tabular}
      
	\end{small}
	\end{center}

\end{table*}

\textbf{Training:} During the experiments stage, we process a set of 2048 uniform sampled points into the network and for each stage graph nearest neighbor value k is constant to 40 throughout the training/testing. To compute the network losses, we employed categorical cross-entropy loss function, and for network optimization we chose the Stochastic Gradient Descent(SGD) approach with a learning rate of 0.01. Like the classification network, we employed a cosine annealing learning rate scheduler for enhanced optimization during training. The batch size was configured to 32, and labels were also pass into the network during the training phase.

\begin{figure}[!b]
    \centering
    {\includegraphics[width=12cm]{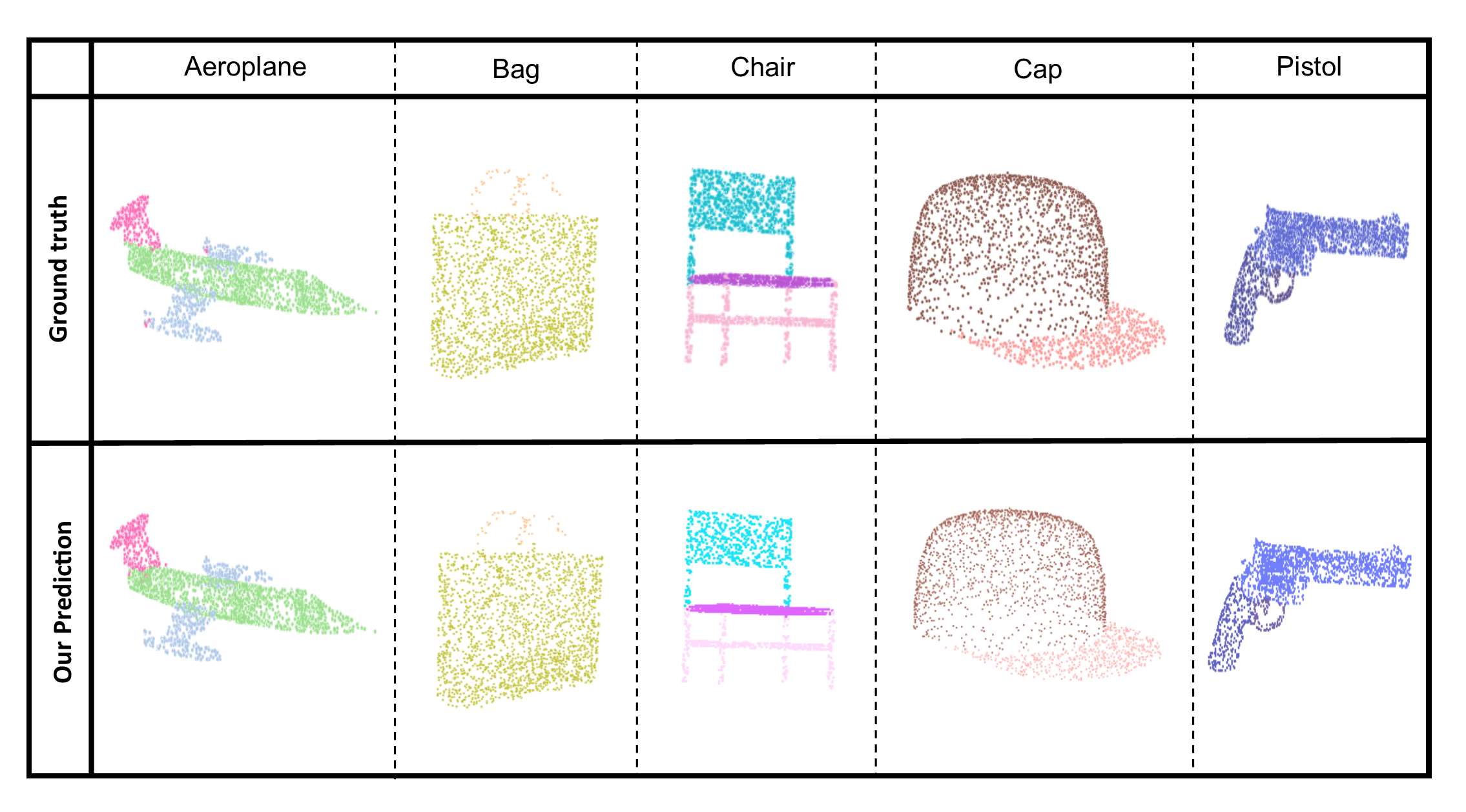} }
    	\caption{ The visual representation shows how the network segments each shape into different parts based on the point cloud input using the open-source Meshlab visualizer. The figure uses a grid layout with two rows: the top row shows the ground truth with multi-colored point clouds, and the bottom row shows the prediction with variations in point cloud colors.}
    \label{fig1}
\end{figure}

\textbf{Results:} The outcomes obtained from the ShapeNet-Part dataset are presented in Table \ref{part-seg-results} and also compared with state of the art result. Our proposed model for part-segmentation task achieved similar result compared to existing networks, the mIoU(mean intersection over union) value beat most of the state of the art result PointNet, Kd-Net etc . Compared to the baseline DC-GNN network, our model notably outperformed by approximately 2\%, 0.6\% and 9 \% in the significant classes of bag, earphones and rocket respectively. It has performed comparable in case of mug class. It has particularly excelled in one individual class i.e. bag out of 16 classes. The visual results of Part-Segmentation model are shown in Fig.\ref{fig1}

\subsection{Indoor Scene Segmentation} 

Here we evaluate our Point-GR Indoor Scene segmentation architecture using the S3DIS(Stanford Large-Scale 3D Indoor Spaces) dataset. This dataset contains 3D scan point clouds covering 6 indoor areas, amounting to a total of 272 rooms. Each sample point is categorized into one of 13 semantic classes, such as board, bookcase, chair, ceiling, and beam, alongside clutter etc. Before passing the dataset into the model, we first pre-process S3DIS dataset according to benchmark standard \cite{qi2017pointnet}. This process divides each room into 1m x 1m blocks and each block contains fixed number of points which is sampled from original dataset and each points has total 9 attributes(XYZ, RGB, and normalized spatial coordinates). For evaluation purpose we have used 6-fold cross validation technique. During the experiments, we fed 4096 uniform sets of sampled points within the network. Here input contains total 9 channels x, y, z , RGB color information, intensity, normal vector


\begin{table}[]
\caption{ The Results of the Indoor Scene Segmentation on the S3DIS dataset is given in the table. }
 \label{S3DIS-results}
\begin{center} 
\begin{tabular}{|l|c|c|}
\hline
\textbf{\textbf{Models}} & \textbf{Mean IoU} & \textbf{Overall IoU} \\ 
\hline PointNet \cite{qi2017pointnet} & 47.6 & 78.5 \\
\hline PoINTCNN \cite{li2018pointcnn} & 65.39 & - \\
\hline DG-CNN \cite{wang2019dynamic} & 56.1 & 84.1 \\
\hline \textbf{Ours} & \textbf{73.47} & 82.79 \\
\hline
\end{tabular}
\end{center}
\end{table}
\textbf{Results:}
The outcomes obtained from the ShapeNet-Part dataset are presented in Table \ref{S3DIS-results}.
Our model demonstrates superior performance with higher mean compared to baseline DG-CNN network. It has outperformed the baseline model with higher mean approx. to 30\%. The proposed network shows comparable result in IoU(Intersection over union) evaluation. 


\subsection{Ablation Study}

In the proposed network we have investigated the various parameters as mentioned in this section. The point transformation network has been used a foundational block in all the three networks for classification, part-segmentation and indoor scene-segmentation problems. 

\begin{table}[!b]
\caption{Comparative Analysis of Model trainable  Parameter for Different task, where $M$ denotes millions in this table.}
\label{model-params}
\begin{center}
\begin{tabular}{|l|c|c|c|}
\hline \textbf{Model} & \textbf{Classification} & \textbf{Part-Seg} & \textbf{S3DIS} \\
\hline Point-Net \cite{qi2017pointnet} & 0.69M & & \\
\hline DG-CNN \cite{wang2019dynamic} & 2.61M & 1.45M & 0.98M \\
\hline \textbf{Ours} & 1.80M & 1.04M & 1.00M \\
\hline
\end{tabular}
\end{center}
\end{table}

\begin{table}[!b]
\caption{Performance analysis of the Point-GR Classification Model Vs Total Number of Points.}
\setlength{\tabcolsep}{0.1em}
\label{model-class-ablation}
\begin{center}
\begin{tabular}{|c|c|c|}
\hline
\textbf{Total Points} & \textbf{Overall Accuracy} & \textbf{Mean Accuracy} \\ 
\hline 2048 & 89.87 & 86.02 \\
\hline 1024 & 92.7 & 89.09 \\
\hline 921 & 92.26 & 89.92 \\
\hline 819 & 92.42 & 89.3 \\
\hline 716 & 91.65 & 87.68 \\
\hline 614 & 90.11 & 85.95 \\
\hline 512 & 88.08 & 83.5 \\
\hline 204 & 47.97 & 40.16 \\
\hline
\end{tabular}
\end{center}
\end{table}

In this part, we examine the different parameters of our proposed work Point-GR architecture in depth. We use Point-GR Transformation transformation network which forms the basis for both object classification and segmentation(part and indoor) use cases.  The number of network parameters generated has been deeply analysed and tabulated in the Table \ref{model-params}. As shown in tabulated result the number of network parameters is significantly reduced in case of classification and part-segmentation network while for S3DIS it is comparable to the baseline model DG-CNN. Overall the network accuracy is similar to baseline DG-CNN model with lesser parameters hence network runtime improved significantly.

Further, the classification network is analysed with with respect to total number of points and accuracy(Overall and Mean accuracy). The ablation studies are performed on ModelNet-40 dataset and the respective results are documented in Table \ref{model-class-ablation}. The model achieved highest overall accuracy on 1,024 number of points, if the number of points increases or decreases the accuracy also degrades. When the number of points is half of total number of points(1,024) the overall accuracy's is shrinked by 4.9 \%. The overall accuracy of the model is not directly proportional to the number of points.

\begin{table}[]
\caption{ Comparative evaluation of Classification network on varied values of Nearest Neighbors($k$=5,10,20...40) points on ModelNet-40 dataset.}
\begin{center} 
\label{k-model-performance}
\begin{tabular}{|c|c|c|}
\hline
\textbf{$\textbf{k}$} & \textbf{Overall Accuracy} & \textbf{Mean Accuracy}\\
    \hline 5 & 29.53 & 24.79 \\
    \hline 10 & 85.21 & 79.26 \\
    \hline 15 & 91.73 & 88.28 \\
    \hline 20 & 92.7 & 89.09 \\
    \hline 30 & 91.12 & 87.23 \\
    \hline 40 & 87.72 & 83.23 \\
    \hline

\end{tabular}
\end{center}
\end{table}

The Point-GR classification model performance is studied and tested with respect to varying value of $k$ i.e total number of nearest neighbors and the obtained results are tabulated in Table \ref{k-model-performance}. For this study we have used a fixed of 1,024 for the overall number of points. As shown in tabulated result, when we increase number of $k$ the overall accuracy and mean accuracy also increases up to $k$ = 20. Once $k$ is increased, there is no subsequent improvement in the model's performance and model achieved highest accuracy when $k$ = 20. This study concludes that as the value of $k$ increases, there is a corresponding rise in accuracy. However, beyond a certain point, further increases in the value of k do not lead to improvements in the model's performance.

Our studies illustrates that the proposed model performs comparable to the baseline model DG-CNN \cite{wang2019dynamic} while utilizing fewer model parameters and exhibiting lower prediction time complexity.

\section{Conclusion}

In this research, we introduced a novel architecture named Point-GR for 3D Point Cloud Object Classification, Part-Segmentation, and Scene Segmentation. The proposed model achieved better results with respect to  prior 3D Scene Segmentation architectures. All the inputs in all the networks are first transformed using Point-GR Transformation network for a greater representation of the input. Through the experiment, we also showed that our backbone network Point-GR Feature Learning Network(F.L.N) has capability to be extended to implement 3D Part-Segmentation and Scene-Segmentation. This proposed work can be further extended on graph construction techniques, edge features and graph learning algorithms. In other directions introducing attention mechanism in feature learning stage. 

The Point-GR model, a deep learning network, can be seamlessly integrated into ADAS use cases like 3D Object Detection, Instance Segmentation, Depth Estimation, and SLAM, as well as in industrial robotics, drone mapping, and surveillance. Its effective feature learning for point cloud data makes it suitable for various research-oriented real-world problems. According to published results, it is the first graph-based lightweight residual model for 3D perception tasks such as Classification, Part-Segmentation, and Scene Segmentation.

\renewcommand{\bibname}{References} 

\bibliography{sn-bibliography}




\end{document}